\documentclass[sigconf]{acmart}

\usepackage{subfigure}
\usepackage{algorithm} 
\usepackage{algorithmic} 
\usepackage{multirow} 
\usepackage{amsmath} 
\usepackage{xcolor}
\usepackage{booktabs}

\AtBeginDocument{%
  \providecommand\BibTeX{{%
    \normalfont B\kern-0.5em{\scshape i\kern-0.25em b}\kern-0.8em\TeX}}}

\setcopyright{acmcopyright}
\copyrightyear{2002}
\acmYear{2023}
\acmDOI{XXXXXXX.XXXXXXX}

\acmConference[2023]{}
{}{}
\acmPrice{}
\acmISBN{}

\begin{document}

\title{Interpreting Hidden Semantics in the Intermediate Layers of 
\\ 3D Point Cloud Classification Neural Network}

\author{Weiquan Liu}
\email{wqliu@xmu.edu.cn}
\affiliation{
   \institution{Xiamen University}
   \city{Xiamen}
   \country{China}
   \postcode{361005}
 }

\author{Minghao Liu}
\email{23020191153186@stu.xmu.edu.cn}
\affiliation{
	\institution{Xiamen University}
	\city{Xiamen}
	\country{China}
	\postcode{361005}
}

\author{Shijun Zheng}
\email{zhengshijun@stu.xmu.edu.cn}
\affiliation{
   \institution{Xiamen University}
   \city{Xiamen}
   \country{China}
   \postcode{361005}
 }

\author{Cheng Wang*}
\email{cwang@xmu.edu.cn}
\affiliation{
   \institution{Xiamen University}
   \city{Xiamen}
   \country{China}
   \postcode{361005}
 }







\renewcommand{\shortauthors}{}

\begin{abstract}
Although 3D point cloud classification neural network models have been widely used, the in-depth interpretation of the activation of the neurons and layers is still a challenge. We propose a novel approach, named Relevance Flow, to interpret the hidden semantics of 3D point cloud classification neural networks. It delivers the class Relevance to the activated neurons in the intermediate layers in a back-propagation manner, and associates the activation of neurons with the input points to visualize the hidden semantics of each layer. Specially, we reveal that the 3D point cloud classification neural network has learned the plane-level and part-level hidden semantics in the intermediate layers, and utilize the normal and IoU to evaluate the consistency of both levels' hidden semantics. Besides, by using the hidden semantics, we generate the adversarial attack samples to attack 3D point cloud classifiers. Experiments show that our proposed method reveals the hidden semantics of the 3D point cloud classification neural network on ModelNet40 and ShapeNet, which can be used for the unsupervised point cloud part segmentation without labels and attacking the 3D point cloud classifiers.
\end{abstract}





\maketitle

\section{Introduction}
Nowadays, deep neural network (DNN) models have outstanding performance in various tasks, especially classification problems. However, the classifier acts as a black box, and it is not clear what the black box learns inside. 
Containing obscure features with high dimensions, it is difficult to understand the decision basis and process of 3D point cloud DNNs.
The obscurity impedes the manipulation and promotion of DNNs, so interpreting what the DNNs learn inside has become an important research field. 

Currently, interpreting the hidden semantics of DNNs give a deeper insight into such a black box. As an intuitive, effective and comprehensive interpretation, hidden semantics shows the activation of neurons or layers inside the network, and associate the activation with the entities, such as lines, surfaces or parts of the input instance. The explicit hidden semantics clarifies the meaning of each neuron and layer, which helps to strengthen the trust of human beings for the network\cite{burkart2021survey}, clarify the decision-making path of the network \cite{cui2020feature}, diagnose the disentangled representation of the network \cite{zhang2018examining}, and promote the researches in other fields, such as transfer learning \cite{yosinski2014transferable}, semi-supervised learning \cite{tritrong2021repurposing}.

\begin{figure}[t]
\begin{center}
        \includegraphics[width=8.5cm,height=5cm,clip]{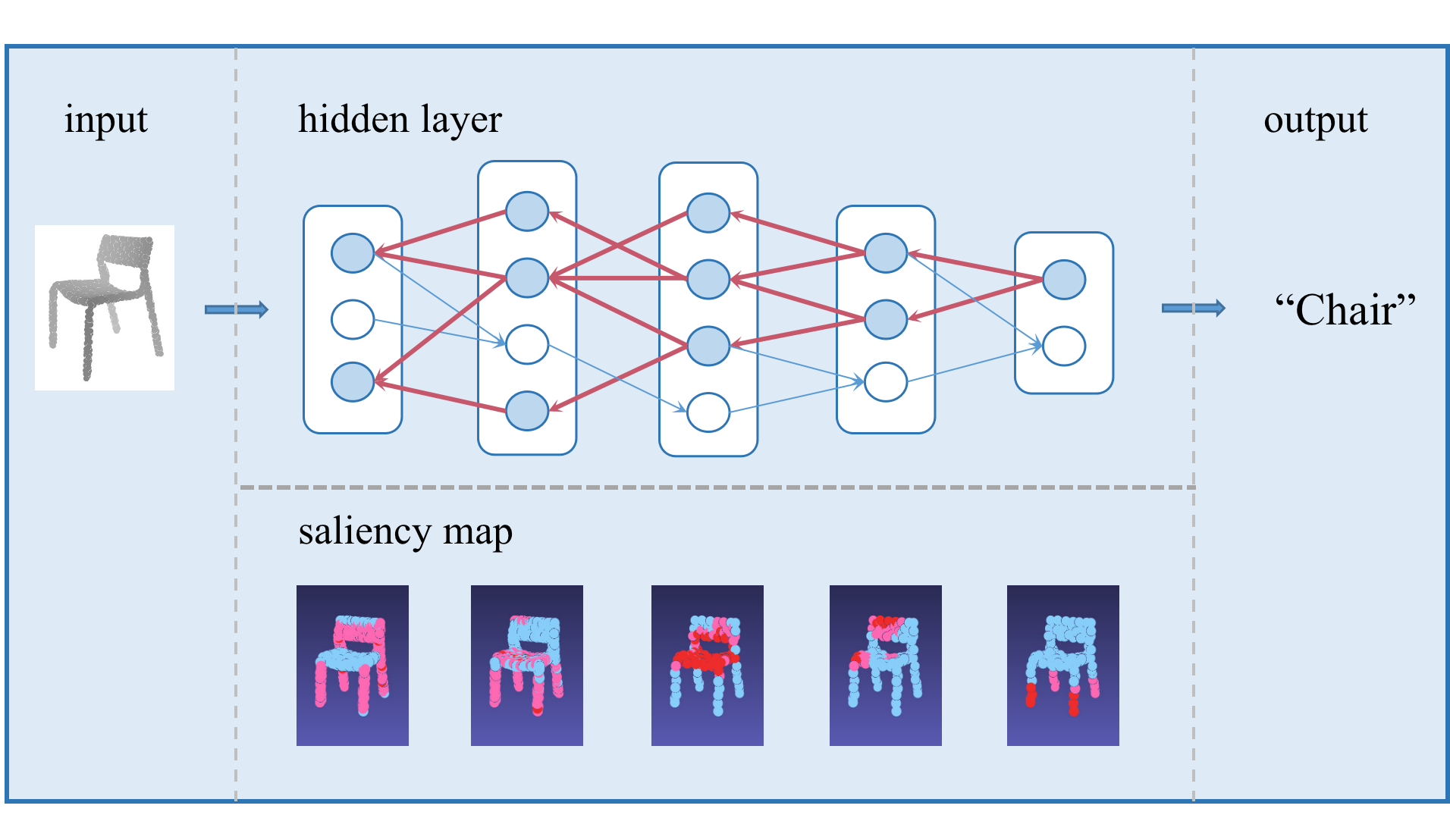}
        \caption{Interpretation of the hidden semantics in 3D point cloud classifier by our method Relevance Flow. Associating the activation of the hidden layer with the 3D input instance, the hidden semantics of each intermediate layer can be seen with the saliency map. Upper: the activation of the hidden layer. Lower: the saliency map of interpreted hidden semantics, in which the red is the most salient region, followed by pink, and blue is insignificant region.}	
        \label{pic_intro}
\end{center} 
\end{figure}

Interpreting hidden semantics is associating the abstract concepts with the activation of some hidden neurons or layers \cite{zhang2021survey}, driven from grandmother cell hypothesis\cite{gross2002genealogy}. It hypothesizes that the grandmother cell, "a hypothetical neuron that represents a complex but specific concept or object"  is activated when a person "sees, hears, or otherwise sensibly discriminates" \cite{clark2000theory} a specific entity, such as their grandmother. Similarly, the interpretation of hidden semantics is to build the relation between the activation of neurons or layers and some kind of entities. The most direct way to interpret the hidden semantics is by visualizing what the neuron or layer is "looking for" with a saliency map, as shown in Figure \ref{pic_intro}.

Although there is some interpretable works on 2D image classifier, it lacks complete research on 3D point cloud classifier. \cite{gupta20203d} showed the corner points contribute more to the classification output, but ignored the interpretation for the intermediate layers, which is the key to open the network black box and represent the learning process of each layer. Besides, current researches are impeded by the unsystematic evaluation criteria for interpretability. \cite{zhao2020evaluation} evaluated the interpretability on internal and external consistency, however, both of them only focus on the max-pooling layer, ignoring a large number of convolutional layers, which is crucial for interpretation. In this case, we study on the interpretation and evaluation of hidden semantics in 3D point cloud classifier intermediate layers.

In this paper, we propose Relevance Flow to interpret the hidden semantics of each intermediate layer. Inspired by LRP \cite{bach2015pixel} and FLOWN \cite{cui2020feature}, we construct a path from output to the input, allowing the Relevance, a value decomposed from the prediction, flows backward, so each activated neurons related to the prediction has a Relevance value. We firstly group the neurons with Relevance value and associate the grouped neurons in same layer to the input points, then visualize the result of each intermediate layer using saliency map. 

Specifically, from the saliency map, we find the plane-level and part-level salient region of hidden semantics appear at intermediate layer in 3D point cloud classifier consistently. We propose utilizing the variance of the salient points normals and the point intersection over union (IoU) of the salient points to evaluate the consistency on the intermediate layers.

In addition, to represent the application of 3D point cloud hidden semantics, we part the segmentation of 3D point cloud sample in an unsupervised manner. We train the 3D point cloud classifiers with class labels only, and use the consistent plane and part level hidden semantics to construct part detector and realize part segmentation. Besides, we generate the adversarial samples by moving the salient regions of hidden semantics to attack 3D point cloud classifiers.

Experimental results show that our proposed method reveals the the hidden semantics of intermediate layer in 3D point cloud classifier. We conduct the experiments with PointNet\cite{qi2017pointnet}, PointNet++\cite{qi2017pointnet++}, and PointConv\cite{wu2019pointconv} framework on the ModelNet10\cite{wu20153d} and ModelNet40 \cite{wu20153d} datasets. We also evaluate both plane-level and part-level hidden semantics of the intermediate layers in PointNet++ on ModelNet40 and ShapeNet\cite{yi2016scalable} datasets. 
Besides, experiments show that our method achieves the 71.3\% mIoU on the ShapeNet part dataset, and  the attack success rate of PointNet is the highest, followed by PointConv and PointNet++.

Our contributions are as follows: 
\begin{itemize}
\item We propose Relevance Flow to interpret the hidden semantics of the intermediate layer in 3D point cloud classification neural networks.
\item We propose the plane-level and part-level criteria to evaluate the consistency of the hidden semantics, which qualitatively descripts the interpretability of the 3D point cloud classification neural networks.
\item Experiments show that the interpreted hidden semantics can be employed for unsupervised part segmentation task and adversarial sample generation task.
\end{itemize}

\section{Related Work} 
\subsection{Interpretation of 2D image classifiers}
\indent The interpretation of 2D image classifiers usually based on attribution, hidden semantics and decision. 

{\bf Attribution} studies the impact of the input on the prediction, and uses a saliency map, a mask of the same size of the input image, to visualize the contribution value of the pixels. ~\cite{springenberg2014striving,selvaraju2017grad,sundararajan2017axiomatic} used the gradient value as the basis of the score, and back propagate it from the output. ~\cite{bach2015pixel,shrikumar2017learning} transformed the gradient value to get better performance. These type of methods are easy for people to understand, but lack of the analysis on the network itself.

{\bf Hidden Semantics} aims to explore the activation of the hidden neurons and layers. To explore the hidden semantics in 2D image classifiers, the most direct way is visualizing the semantics appeared in the neurons by training the input image. \cite{simonyan2014deep,zeiler2014visualizing,mahendran2015understanding} visualized the appearance that maximized the activation of the given neurons. \cite{olah2017feature} refined the appearance into edges, textures, patterns, parts and objects. Due to the severe deformation, the appearance is unrecognizable for human beings. Instead of training the input images, \cite{wang2018visualizing} preserved readability in some way by blurring the input image, but it is still not intuitive. \cite{zhang2018interpretable} is an activation method to show the explicit semantics by masking the filters, and \cite{gonzalez2018semantic} gave the application about the semantics. However, both of them rely on the interpretability of the original network. 

{\bf Decision} represents the decision path of the 2D image classifiers, usually by graph\cite{zhang2018interpreting} or decision tree \cite{zhang2019interpreting}. \cite{cui2020feature} proposed the method visualizing the saliency map of each layers by constructing an activation path from the output to the input, and constructed the decision path. Compared with other methods based on decision, \cite{cui2020feature} is relatively simple, and shows the semantics of the hidden layers, which is easy to understand for human beings. Inspired by it, we propose the Relevance Flow to interpret the hidden semantics of the 3D point cloud classifiers.

\subsection{Interpretation for 3D point cloud classifiers}
\indent At present, there are few works interpreting the 3D point cloud classification neural networks. \cite{gupta20203d} analyzed the sensitivity of the network to the input 3D point cloud based on the gradient value. It points out that the network pays more attention to corner points, but the conclusion is only at the input level and lacks inter-layer research. Although \cite{zhao2020evaluation} defined the inter and outer consistency for 3D point cloud classification neural network, but it lacks an in-depth interpretation to answer what have been learned in the intermediate layers and how the network makes its decision. In this  paper, we explore the answer by interpreting the hidden semantics in the intermediate layers.

In addition, to test the vulnerability of 3D point cloud classifiers, PointGuard\cite{liu2021pointguard} attacked the network by adding, deleting, and modifying the input points. However, it requires a lot of calculations on looking for the modified points, and the principle is relatively complicated, which is not suitable for real application. \cite{zheng2019pointcloud} used Shaply Value to study the sensitivity of the network to the input regions, which is more intuitive and concise than PointGuard, but it mainly focuses on the single sample, lack of quantified research on the class-level, such as which category is more likely to be attacked.

\section{Method}
\subsection{Overview}
\indent  Hidden semantics is designed to associate abstract concepts with the activation of some hidden neurons \cite{zhang2021survey}, described in mathematical language as
\begin{equation}
r: act(x; \theta) \sim x  \label{act1} 
\end{equation}
\noindent where $x$ is the input, and $act(x; \theta)$ represents the activation of the neurons or layers in the network. $\sim$ denotes the association between the activation and input, and $r$ means the consistency of the association.

\begin{figure}[!t]
\begin{center}
        \includegraphics[width=\linewidth]{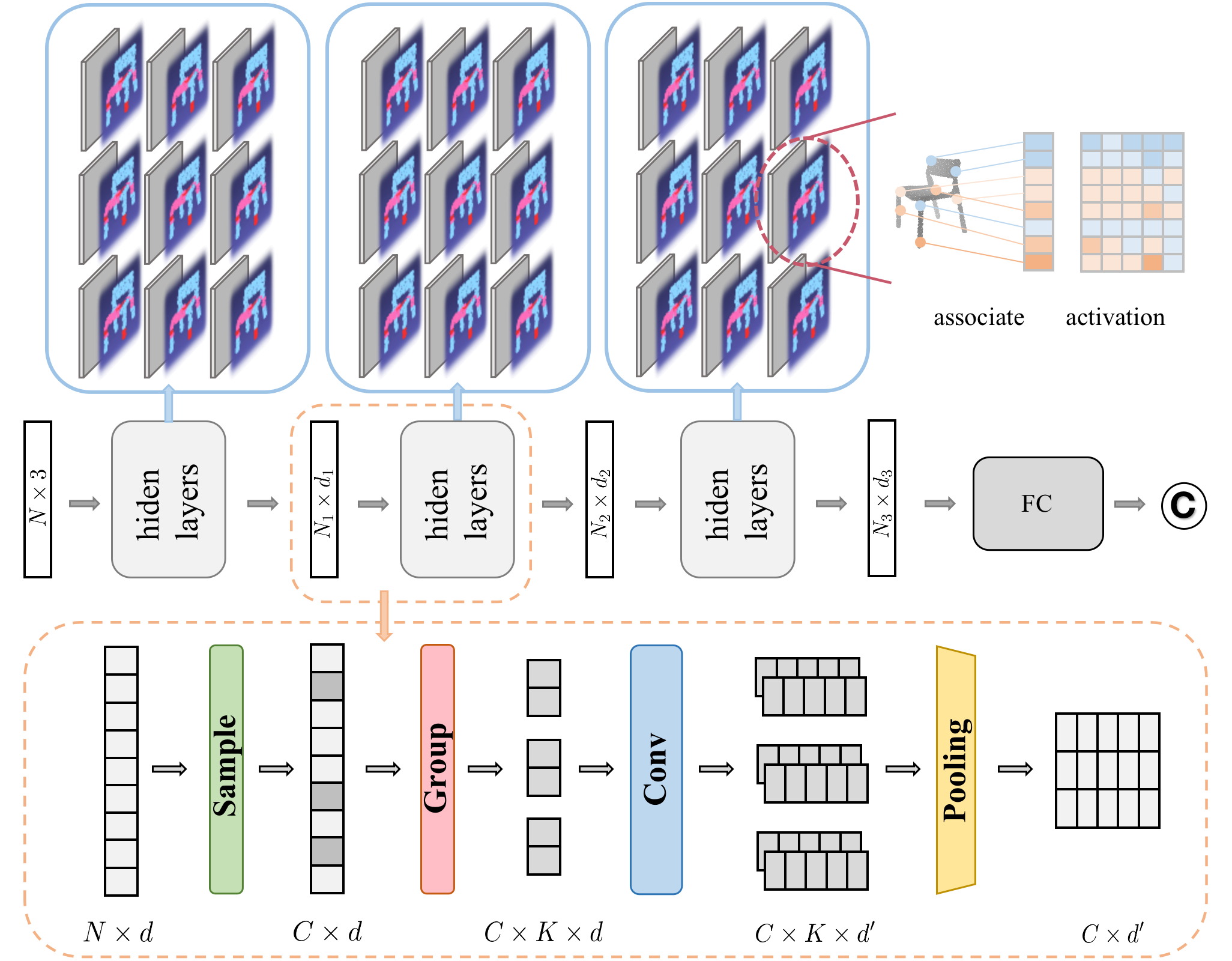}
        \caption{Upper: Manifestation of the hidden semantics in the intermediate layers. Lower: The framework of the 3D point cloud classifier.}	
        \label{method0}
\end{center} 
\end{figure}

\indent  We manifest the interpreted hidden semantics of each intermediate layer of 3D point cloud classifier by saliency map, as shown in Figure \ref{method0}.

{\bf Sample} operation samples the input data as $X_{center} = s(X)$, where the input data of size $N \times d$ denoted as $X=\{x_1, x_2, \cdots, x_n\} \in R^N$, and $X_{center}$ means the sampled data with the size of $C \times d$, where $C$ represents the number of sampling. {\bf Group} operation combines the center and its neighbors, so that the data is divided into several local regions. It conducts as $X_{c}^{'} = g (X_{center}, X_c)$, where $X_c = \{x_c^1, x_c^2, \cdots, x_c^K\}$, and $K$ is the number of neighbors. For each of the $x_c^k \in X_c$, $g$ conducts as $x_c^k-x_{center}$. {\bf Conv} operation convolves each Group separately as $H_{c} = h (X_c^{'}) $, where the size of $H_{c}$ is $K \times d^{'}$, and $d^{'}$ depends on the depth of the convolution. Moreover, $h$ concerns three operations: conv1D, conv2D, conv3D. {\bf Pooling} operation compress all the data in each group and use less data for representation, following $X^{'} = p (H_{c})$, where $x^{'}$ with the size of $d^{'}$. 

The operations above are concluded as $X^{'} = p ( h ( g(s(X),X) ))$, which conducted by a trained 3D point cloud classifier. Our method is also applicable to other modules of the 3D point cloud neural network, such as T-Net\cite{qi2017pointnet} and Attention module\cite{wang2019graph}. In order to demonstrate the principle of the method, we take the PointNet++ module as an example to illustrate.

\subsection{Relevance Flow}
\begin{figure}[t]
  \centering
  \includegraphics[width=\linewidth]{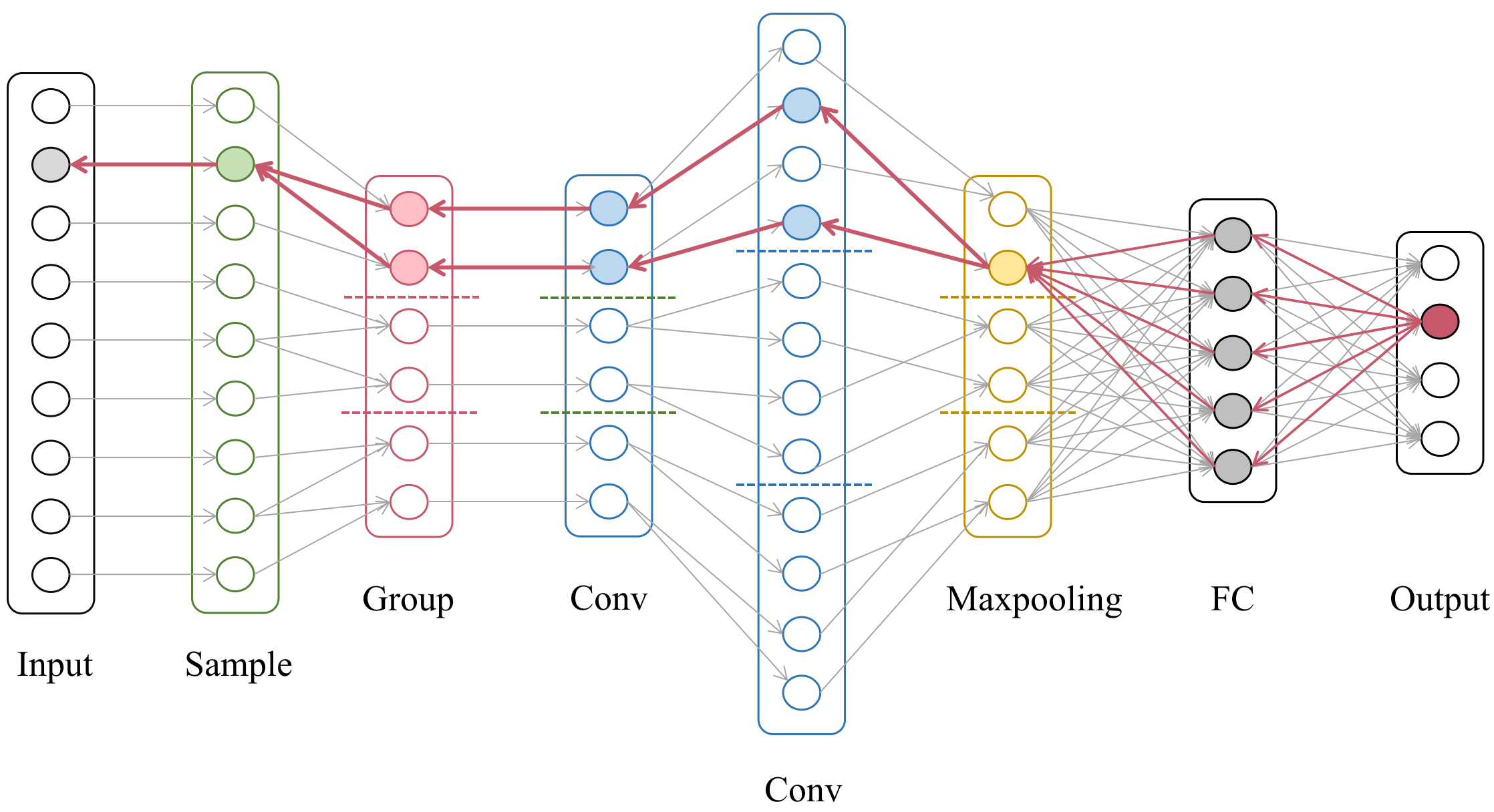}
  \caption{Relevance flow in the 3D point cloud classifier. The Relevance value is propagated from the Logits layer to the Input layer with the activation of the neurons.}
  \label{pic_method2}
\end{figure}

\indent As shown in the Figure \ref{pic_method2}, the path constructed by the red arrow represent a process that Relevance flows from output to input. At first, we decompose the prediction into a vector as the manner of LRP\cite{bach2015pixel} (decomposing the prediction of a deep neural network down to the relevance scores, which is a $[0,1,0,0]$ vector, and only the position where the predicted value is largest with value $1$, the rest are $0$). Then starting from the output layer, the Relevance flows to the activated neurons in the previous layer FC (Fully Connected Layer) as the back propagation manner, but the propagated value is no longer the gradient (The propagated value is calculated with the formula detailed in the following paragraphs). Finally, the activated neurons relevant to the prediction have the Relevance value in FC layer. By analogy, the Relevance flows at input layer, and all activated neurons related to the predict have Relevance value.

 The Relevance value between the activated neurons and output prediction is represented as $R$. If the $R$ value of a neuron is larger, it means that the neuron is more related to the output prediction. The $R$ is calculated with the formula:
\begin{equation}
R_i = \sum_{0}^{j}{ \frac{a_{i}w_{ij}}{\sum_{0}^{i}a_{i}w_{ij}}R_j}  \label{basic_lrp} 
\end{equation}

\noindent where $a_i$ is the activation value of the $neuron_i$, and $w_{ij}$ is the weight of the $neuron_i$ to the $neuron_j$ in next layer. $R_i$ is the relevance value of the neurons in current layer, and $R_j$ is the relevance value of the neurons in next layer. $\sum$ means the current relevance of the activated neurons relies on all the activated neurons related to it. The $w_{ij}$ is further described in the following paragraphs.

 \indent For a trained 3D point cloud classifier, the formulas of Relevance Flow are as follows:

\begin{figure}[t]
\centering
\includegraphics[width=3cm]{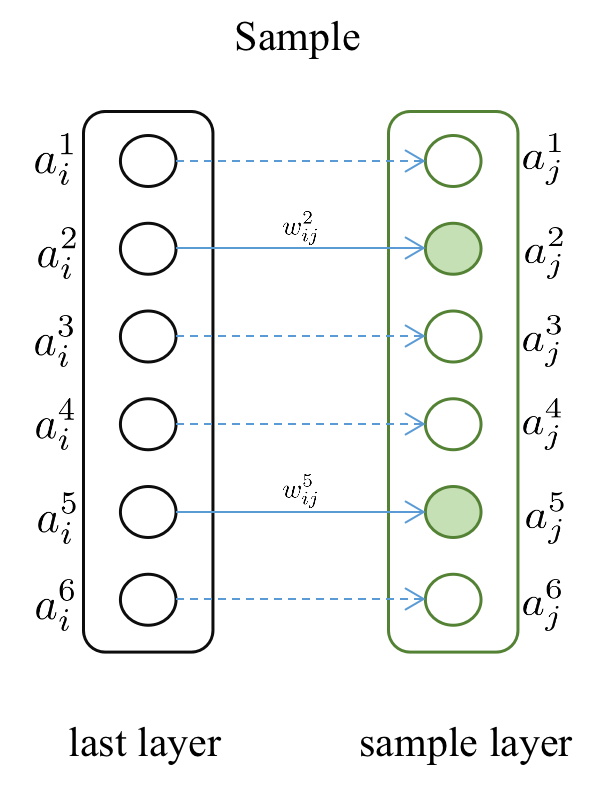}
\quad
\includegraphics[width=3cm]{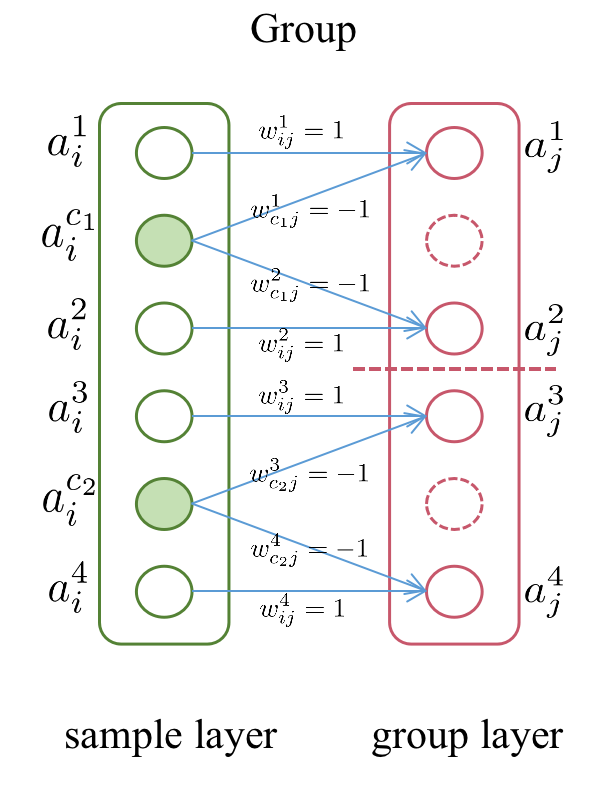}
\caption{The Relevance Flow in sample and group. }
\label{method2_2}
\end{figure}

In {\bf Sample} operation, showing in the Figure \ref{method2_2}, $w_{ij}=1$, so
\begin{equation}
R_{i} = R_{j}  \label{sapmle_lrp} 
\end{equation}

In {\bf Group} operation, the grouped data is related to the value of center and the value of itself before grouping, with the condition $a_{i}-a_{center}=a_{j}$, the weight of $a_i$ to $a_j$ is 1, and the weight of $a_{center}$ to $a_j$ is -1. So the formula is
\begin{equation}
R_{i} = \sum_{0}^{j}\frac{a_{center}}{a_{center}-a_{i}}R_{j}  \label{group_lrp} 
\end{equation}

In {\bf Conv} operation
\begin{equation}
R_i = \sum_{0}^{j}{ \frac{a_{i}w_{ij}}{\sum_{0}^{i}a_{i}w_{ij}}R_j} \label{conv_lrp} 
\end{equation}

{\bf Maxpooling} can be regarded as a fully connected layer, only the pooled data weights 1, and the rest is 0, following
\begin{equation} 
R_{i} = 
\left\{ \begin{aligned} 
0     & , & j \neq index\\ 
R_{j} & , & j = index 
\end{aligned} \right. 
\label{pool_lrp}
\end{equation}

\subsection{Evaluation of Hidden Semantics}
Through the saliency map of the interpreted hidden semantics, we find that there are {\bf plane-level} and {\bf part-level} hidden semantics appeared in the Intermediate layers of the 3D point cloud networks consistently. Therefore, in each of the layers separately, we ultimate the normals and IoU of the salient points to measure the consistence of the plane-level and part-level hidden semantics. 

To evaluate the consistency of the {\bf plane-level} hidden semantics, we ultimate the normals of the salient points in the salient maps of each layer, following
\begin{equation}
C_p = \frac{n_p}{n_t}
 \label{normal} 
\end{equation}
\begin{equation}
n_p = \mathbb{I}_{var(N_i)<\gamma}
\end{equation}
where $n_p$ means the number of qualified objects, which measured by the variance of the salient point normals $N_i$ with a thread $\tau$. $n_t$ is the number of the total object in the same class. $C_p$ represents the score of the plane-level consistency.

To evaluate the consistency of the {\bf part-level} hidden semantics, we take the IoU as the criterion. Different from the segmentation task, we trained the network with class labels, and test the $IoU$ using segmentation labels. The formula is 
\begin{equation}
IoU = \frac{n_{cor}}{n_{cls} + n_{seg} - n_{cor}}
 \label{miou} 
\end{equation}
where $n_{cor}$ means the numbers of points with the correct match. $n_{cls}$ represents the total numbers of the salient points, and $n_{seg}$ represents the numbers of the points in the part segmentation area.

\subsection{Unsupervised Part Segmentation Using Hidden Semantics}
\begin{figure}[t]
\centering
\includegraphics[width=\linewidth]{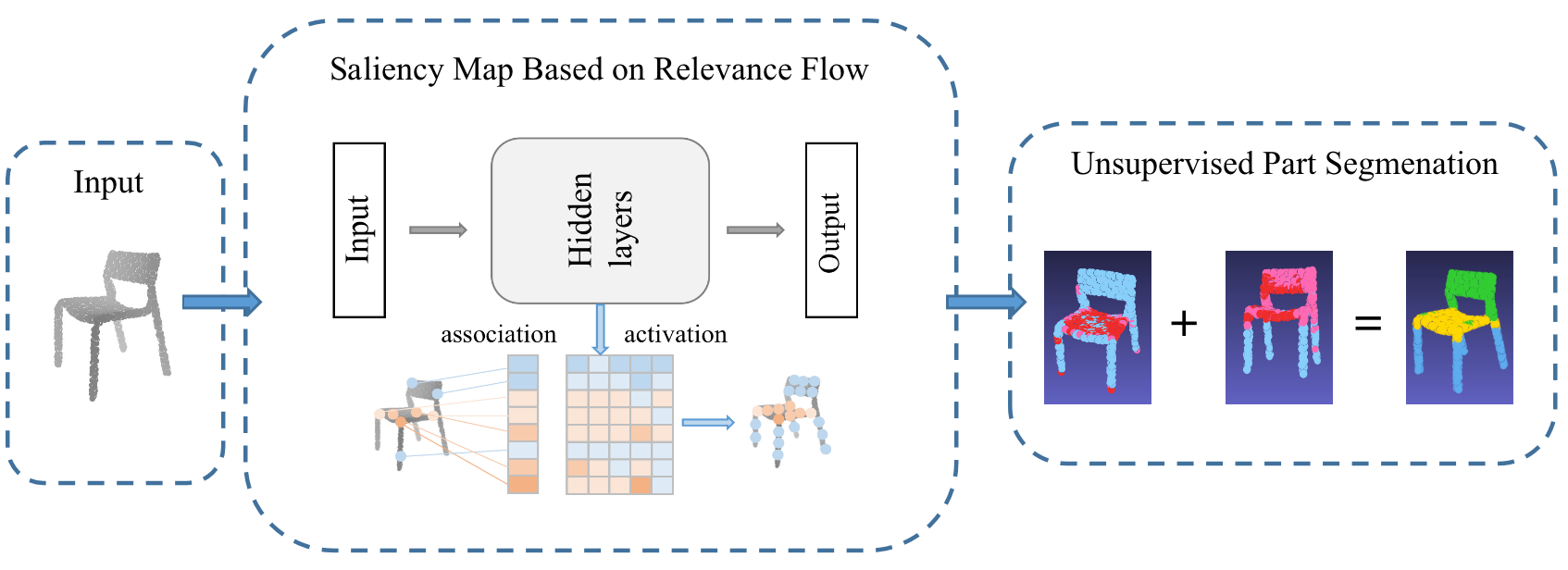}
\caption{Unsupervised part segmentation using hidden semantics. }
\label{meth_seg}
\end{figure}

To reduce the dependence on the label, we implement unsupervised 3D point cloud part segmentation using the hidden semantics, as shown in Figure \ref{meth_seg}.

First, we train the 3D point cloud DNNs with class-level labels. Secondly, we use Relevance Flow to obtain the hidden semantics of the intermediate layers, and select qualified part-level hidden semantics as part detector. Finally, we combine each part detector to realize unsupervised part segmentation of 3D point cloud samples. 

\subsection{Generation of Adversarial Attack Samples}
To explore the robustness and vulnerability of the 3D point cloud DNNs, we generate adversarial attack samples and attack the network, making the output wrong. 

At first, we sort the salient points according to their salient value $w$ derived from Relevance Flow. 
Secondly, we pick up the top $N$ points as the centers, and group their $K$ neighbors to form the attacked regions.
Then, we generate the adversarial attack samples by moving the attacked regions to the non-salient area.
Finally, we input the adversarial attack sample to the network, and obtain the new classification results. If the result is inconsistent with the original category, the attack succeeds, otherwise it fails. The attack progress is shown in Algorithm \ref{alg_attack}.

\begin{algorithm}
\caption{Attacking 3D Point Cloud DNNs} 
\label{alg_attack} 
\begin{algorithmic}[1] 
\REQUIRE $inputdata \in \mathbb{R}^{N \times 1}$, \space	$Network$, \space $N$, \space $K$ %
\ENSURE $success$ \space or \space $fail$
\STATE $cls\_ori = Network.forward(inputdata)$
\STATE $sample = inputdata$
\STATE sort $inputdata$ by $w$
\STATE pick top $N$ points as $sail\_points$
\FOR {$i=1$ to $N$}
\STATE $id = Knn(inputdata,i,K)$
\STATE move $sample_{id}$
\ENDFOR
\STATE $cls\_new = Network.forward(sample)$
\IF {$cls\_new != cls\_ori$}
\RETURN{$success$}
\ELSE
\RETURN{$fail$}
\ENDIF
\end{algorithmic} 
\end{algorithm}

\begin{figure*}[t]
\centering
\subfigure[PointNet]{
\begin{minipage}[b]{0.18\linewidth}
\includegraphics[width=3.5cm,height=7cm]{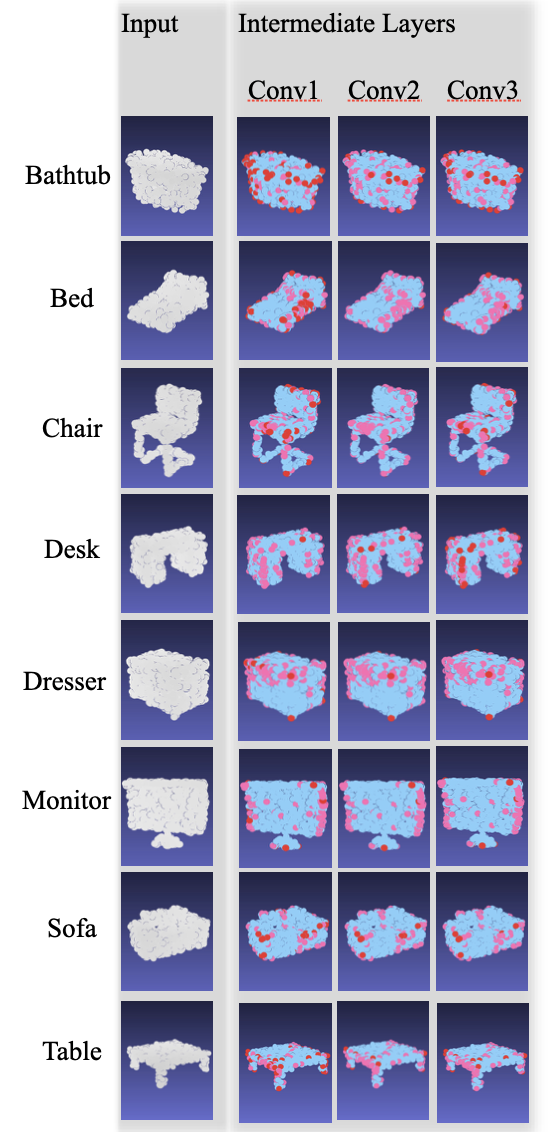}
\end{minipage}}
\subfigure[PointNet++]{
\begin{minipage}[b]{0.38\linewidth}
\includegraphics[width=7cm,height=7cm]{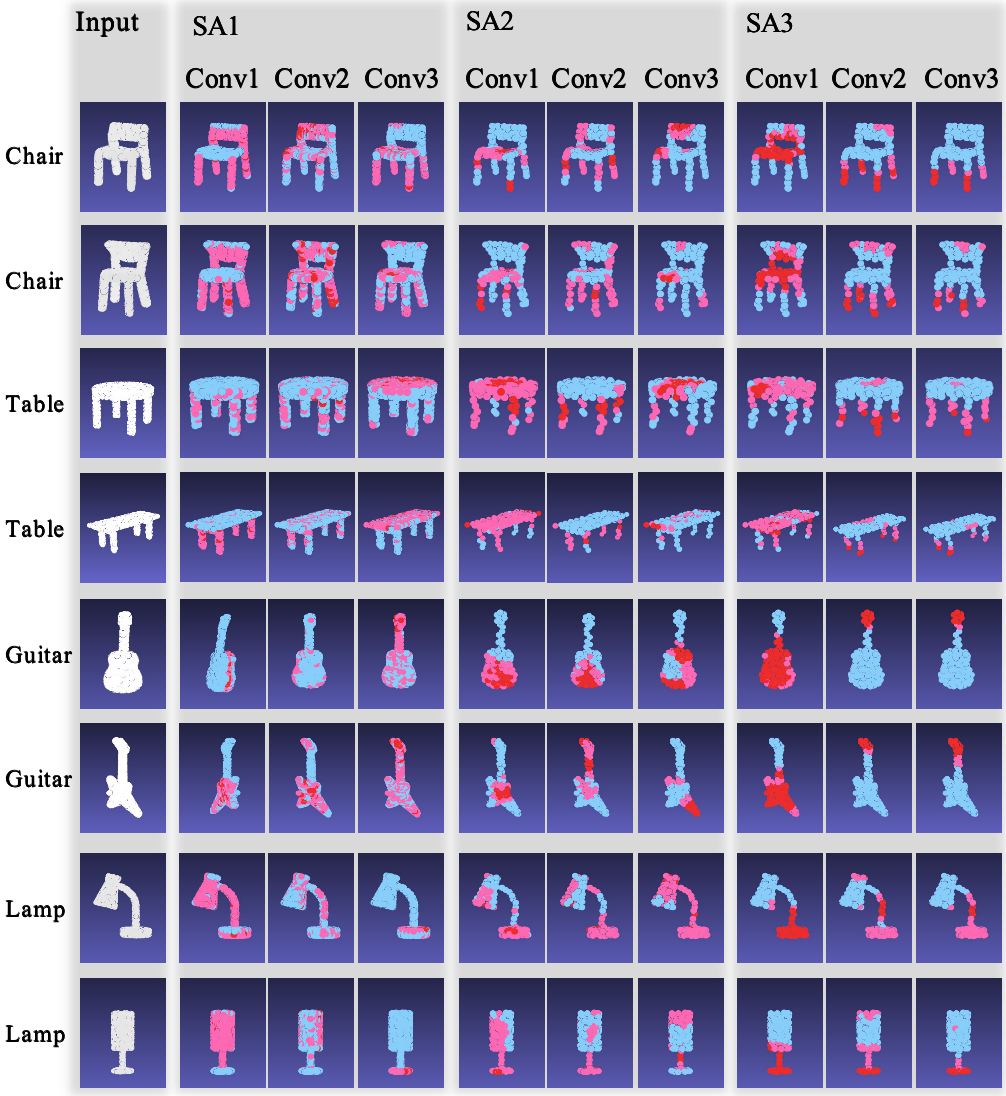}
\end{minipage}}
\subfigure[PointConv]{
\begin{minipage}[b]{0.35\linewidth}
\includegraphics[width=7cm,height=7cm]{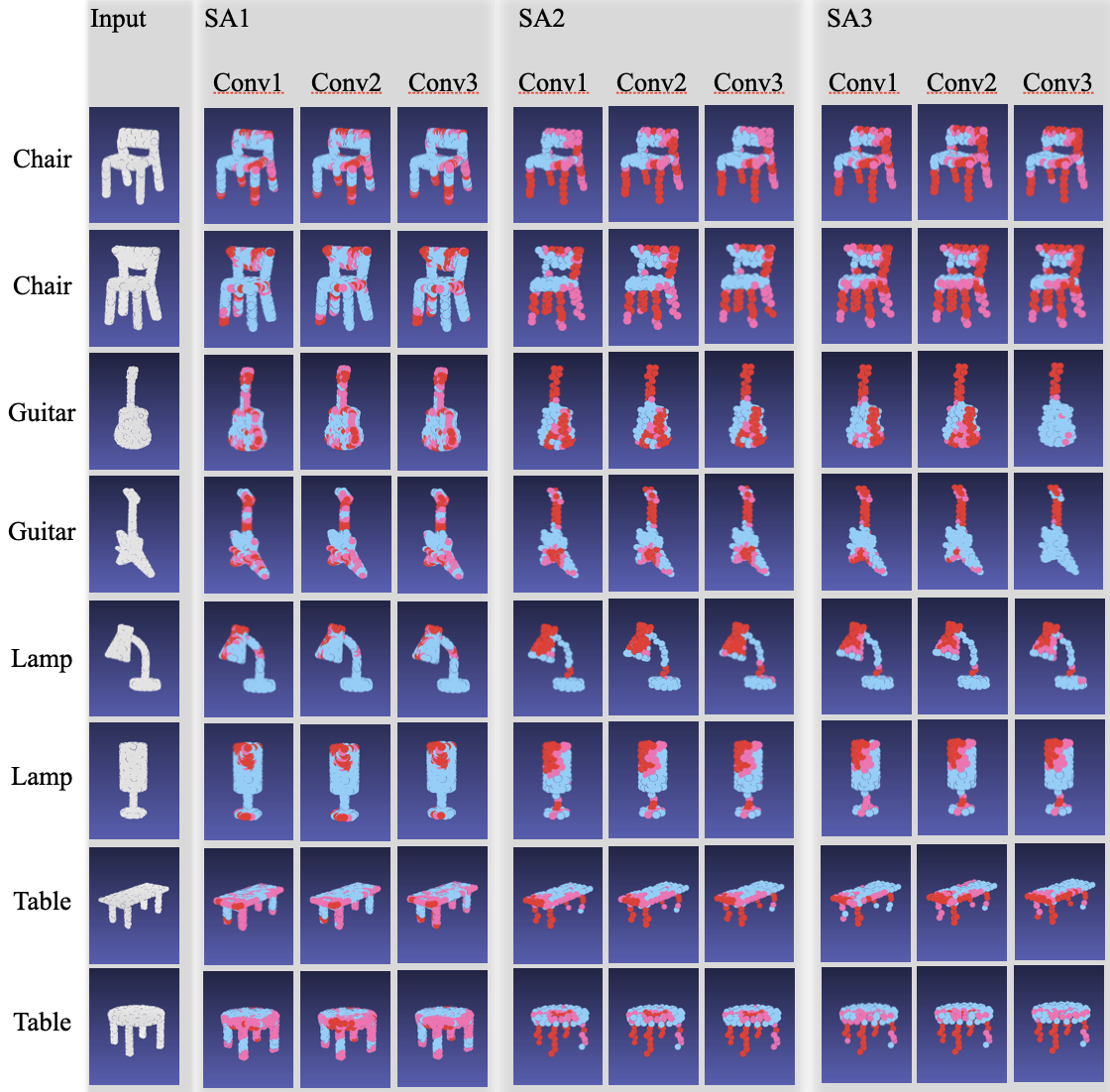}
\end{minipage}}
\caption{The hidden semantics in the intermediate layers of the networks. "SA" represents the sample and group module\cite{qi2017pointnet++} in PointNet++ framwork. The saliency map in each column is corresponding to the hidden semantics of each layer, where from the left to the right is the shallow layer to the deep layer. In each saliency map, the red is salient region, which reflects the concentration of the layer. The red region is the salient region, and the blue region is the insignificant region.}
\label{network}
\end{figure*}

\section{Experiment} 

\subsection{Hidden Semantics of the Intermediate Layer}
In this section, we represent the hidden semantics of PointNet\cite{qi2017pointnet},  PointNet++\cite{qi2017pointnet++} and PointConv\cite{wu2019pointconv} frameworks using our proposed Relevance Flow method on ModelNet10\cite{wu20153d} and ModelNet40\cite{wu20153d} dataset.

The saliency map of the hidden semantics in the intermediate layer is shown in Figure \ref{network}. The salient values are divided into three levels ordered the given interval with $[0, \frac{1}{3}d, \frac{2}{3}d, d]$, where $d$ means the difference between maximum and minimum of the salient value. In Figure \ref{network}, the blue region is insignificant, the pink is more significant, and the red is the most significant. The network is trained strictly as the original setting, due to the space limitation, we only display the hidden semantics of the layers in PointNet++ with radius $0.1$, more details are shown in the supplementary material.  

From the learning process of hidden semantics, PointNet++ focuses on different parts of the instance while the attention become concentrated. From Figure \ref{network}, we can see that the attention of the shallow layer close to the input is scattered, such as the SA1-conv2 (the second convolutional layer in first sample and group module\cite{qi2017pointnet++}). The network learns the surface structure, such as the top plane of table and guitar, as well as the top surface of lamp holder in the SA1-conv3. In the deeper layer, the focus of the classifier on the objects tends to be concentrated, like SA2-conv1 and SA3, focusing on a region, like chair seat, chair leg, table leg, guitar body, and lamp holder. 

However, in Figure \ref{network}, PointNet and PointConv focus on the similar parts of the instance with the progress of strengthening the concentration from shallow layer to deep layer. 
In PointConv network, the attentions of conv1 to conv3 in SA1 module are more scattered, while focused in SA2 and SA3 modules. For example, the class "guitar", the pink and red dots locate at the its body and neck in SA1 module, but in SA2 the number of red dots is increase, and in SA3 module the red dots mainly locate at its neck.

\subsection{Comparison of Salient Region}

\begin{figure}[t]
\begin{center}
        \includegraphics[width=0.45\linewidth]{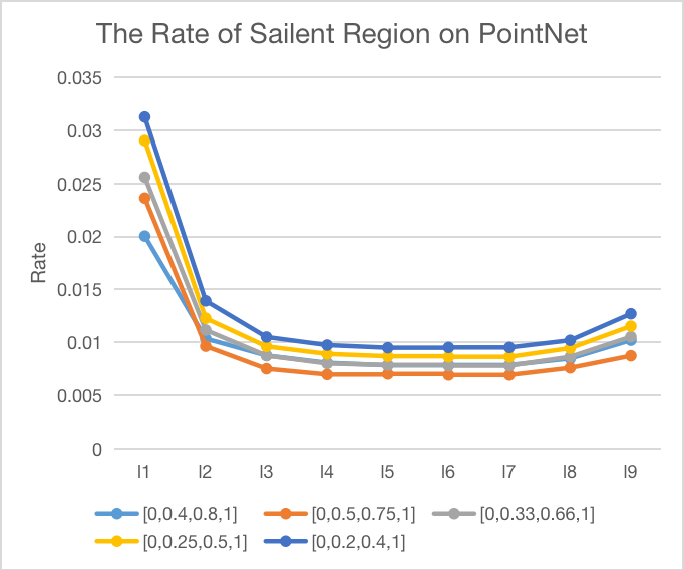}
        \includegraphics[width=0.45\linewidth]{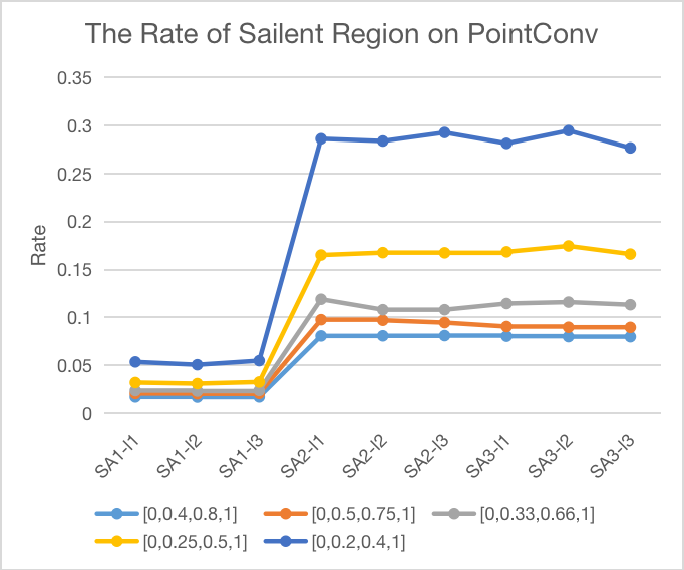}
        \includegraphics[width=0.45\linewidth]{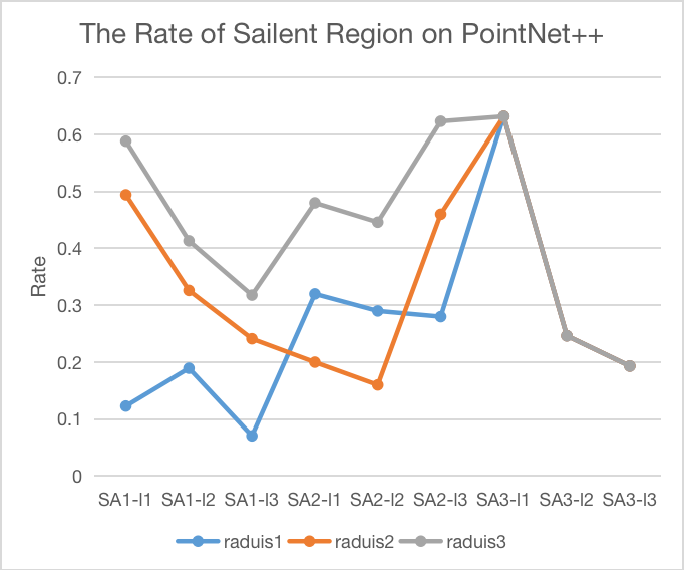}
        \includegraphics[width=0.45\linewidth]{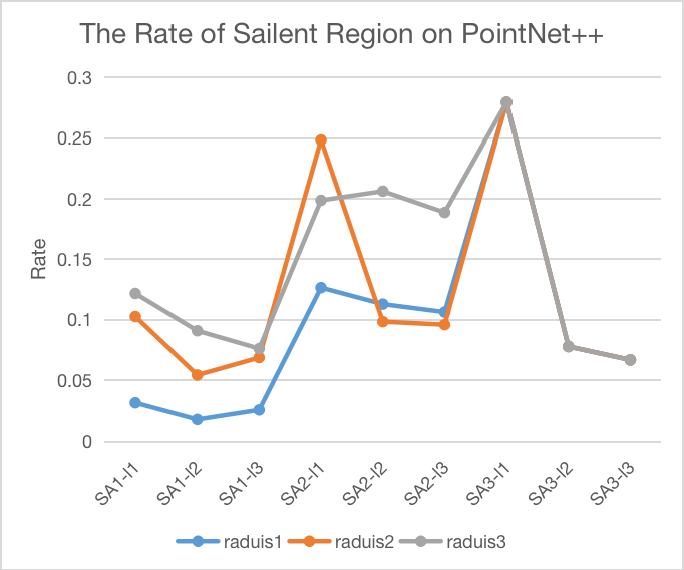}
        \caption{The rate of salient region in the Intermediate Layer.}	
        \label{ex_sal_rate}
\end{center} 
\end{figure}

We compare the trend of the proportion of salient regions with different thresholds on PointNet, PointNet++, and PointConv network, as show in Figure \ref{ex_sal_rate}.

From the aspect of trend, the proportion of PointNet salient region is decreased, and the proportion of PointConv salient region is increased under different thresholds setting. The proportion of PointNet++ salient region fluctuates, especially in the SA3 module, the proportion jumps significantly.

Concerned about the diversity of hidden semantics, PointNet++ has diverse hidden semantic, containing different structures and parts of the instance in different layers, that is the reason we choose this framework to conduct the rest experiments on Section 4.3 and Section 4.4.

\subsection{The Consistency of Hidden Semantics}

Through the experiments, we find the salient regions of different sample in same category appear consistently in some layers.

\begin{figure}th]
\begin{center}
        \includegraphics[width=\linewidth]{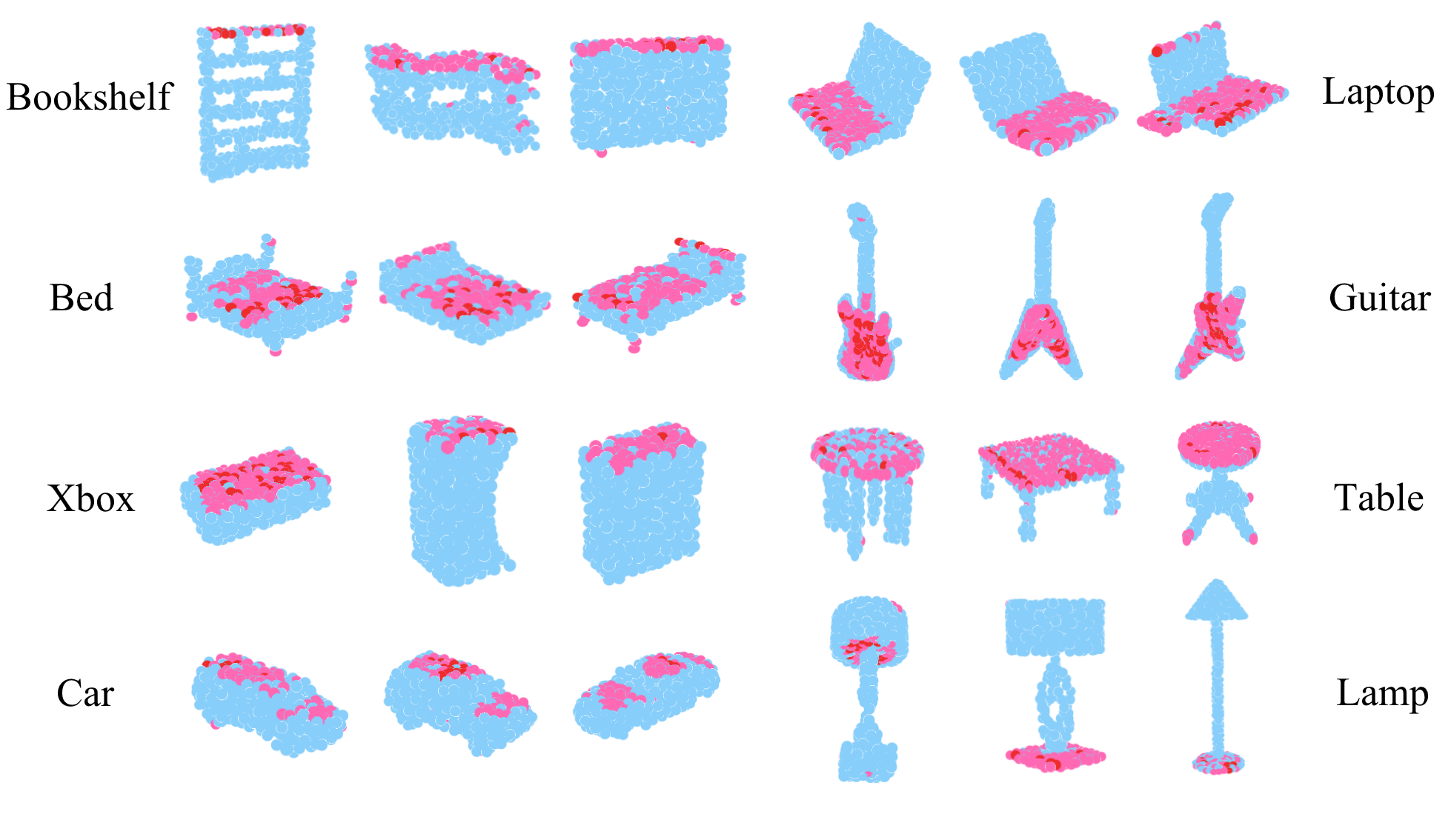}
        \caption{The plane-level hidden semantics in the intermediate layers of PointNet++ on ModelNet40. }	
        \label{ex_plane}
\end{center} 
\end{figure}
\begin{figure}[t]
\begin{center}
        \includegraphics[width=\linewidth]{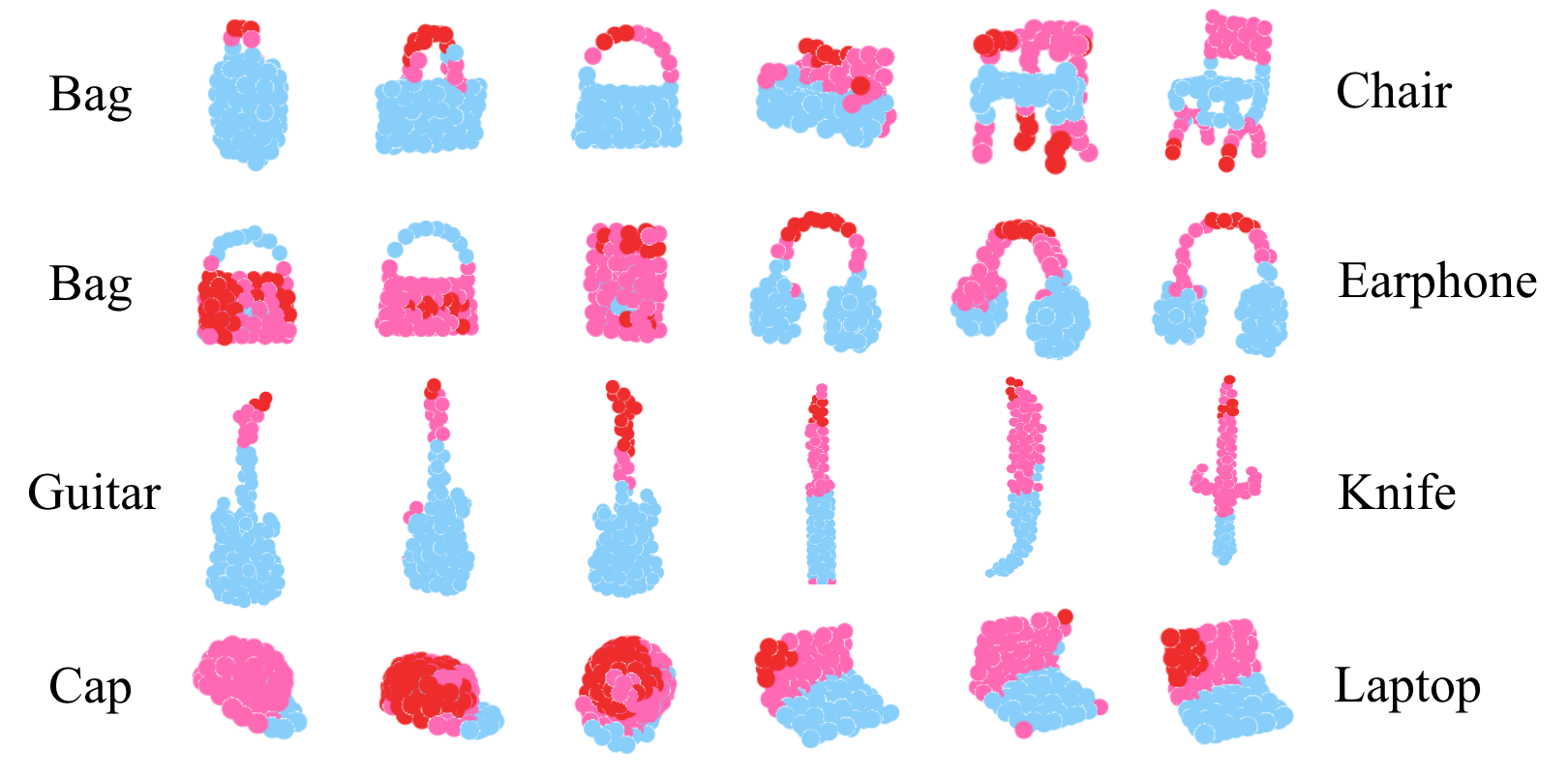}
        \caption{The part-level hidden semantics in the intermediate layers of PointNet++ on ShapeNet. }	
        \label{ex_part1}
\end{center} 
\end{figure}

Especially, we find PointNet++ has learned the plane-level and part-level consistent hidden semantics on ModelNet40, as shown in Figure \ref{ex_plane} and Figure \ref{ex_part1}.

\subsection{Evaluation of Consistency}

Discovering the plane and part level hidden semantics, we evaluates the semantics at both levels separately on ModelNet40\cite{wu20153d} and ShapeNet\cite{yi2016scalable} part datasets.

As shown in Table \ref{tab_plane}, we evaluate the plane-level consistency of PointNet++ on ModelNet40 dataset. We choose the thread $\tau$ as 0.15, which can detect the plane structure with a fine range. Because of the limit of paper, we display the maximum $C_p$ in each of the layer, which mainly comes from the SA1-layer3, and the whole data is detailed in the supplementary material.

\begin{table}[h]
\centering
\caption{The consistency of plane-level hidden semantics of PointNet++ on ModelNet40.}
\begin{tabular}{cccc} 
      \toprule
       Class & $C_p$ & Class & $C_p$\\
       \midrule
       airplane             & 0.80  & laptop               & 0.98\\
       bathtub              & 0.86  & mantel               & 0.99\\
       bed                  & 0.98  & monitor              & 0.67\\
       bench                & 0.97  & night \space stand   & 0.98\\
       bookshelf            & 0.93  & person               & 0.53\\
       bottle               & 0.64  & piano                & 0.71\\
       bowl                 & 0.64  & plant                & 0.53\\
       car                  & 0.88  & radio                & 0.95\\
       chair                & 0.66  & range \space hood    & 0.90\\
       cone                 & 0.78  & sink                 & 0.95\\
       cup                  & 0.70  & sofa                 & 0.88\\
       curtain              & 0.84  & stairs               & 0.58\\
       desk                 & 0.98  & stool                & 0.91\\
       door                 & 0.84  & table                & 0.96\\
       dresser              & 0.97  & tent                 & 0.87\\
       flower \space pot    & 0.48  & toilet               & 0.89\\
       glass \space box     & 0.96  & tv \space stand      & 0.98\\
       guitar               & 0.97  & vase                 & 0.40\\
       keyboard             & 0.99  & wardrobe             & 0.90\\
       lamp                 & 0.88  & xbox                 & 0.96\\
       \bottomrule
    \end{tabular}
\label{tab_plane}
\end{table}

Table \ref{tab_part} shows the part-level hidden semantics of the intermediate layers in PointNet++ framework on  {\em ShapeNet }. We only trained the network with the 16 class labels represented in first row, and test the IoU with the 49 segmentation labels partially represented in second row, detailed in the supplementary material.

\begin{table*}[h]
\centering
\caption{The IoU of part-level hidden semantics of PointNet++ on ShapeNet.}
\begin{tabular}{ccccccccccc}
\toprule
\multirow{2}{*}{Class}  & \multirow{2}{*}{Seg ID}      
& \multicolumn{3}{c}{SA1}
&\multicolumn{3}{c}{SA2}   & \multicolumn{3}{c}{SA3}\\
\cline{3-11}
    &   &  layer1 & layer2 & layer3 & layer1 & layer2 & layer3
        &  layer1 & layer2 & layer3 \\
\midrule
airplane & 0 & 0.55	& 0.54	& 0.56 &  0.43	& 0.33	& 0.53 & 0.53 & 0.46 & 0.36 \\
bag   & 5 & 0.94 & 0.93	& 0.81 & 0.70 & 0.74 & 0.92 & 0.90 & 0.88	& 0.77 \\
cap   & 6 & 0.73 & 0.72	& 0.65 & 0.72 & 0.60 & 0.69 & 0.72 & 0.72 & 0.71\\
car   & 11 & 0.71 & 0.71 & 0.61 & 0.70 & 0.66 & 0.71 & 0.70 & 0.70	& 0.68\\
chair & 13 & 0.47 & 0.48 & 0.44 & 0.45 & 0.36 & 0.57 & 0.48	& 0.37 & 0.30\\
earphone & 16 & 0.63 & 0.63 & 0.57 & 0.53 & 0.50 & 0.64 & 0.60 & 0.54	& 0.14 \\
guitar   & 21 & 0.61 & 0.76 & 0.62 & 0.73 & 0.73 & 0.82 & 0.68 & 0.16 & 0.81 \\
knife    & 22 & 0.50	& 0.51 & 0.53 & 0.65 & 0.52	& 0.54 & 0.48 & 0.53 & 0.65\\
lamp     & 25 & 0.61 & 0.60 & 0.60 & 0.55 & 0.50 & 0.63 & 0.64	& 0.57 & 0.53\\
laptop  & 28 & 0.61 & 0.60 & 0.58 & 0.37 & 0.21 & 0.83 & 0.53 & 0.44 & 0.56\\
motorbike  & 35	& 0.65 & 0.66 & 0.65 & 0.63	& 0.53 & 0.66 & 0.67 & 0.63 & 0.66 \\
mug  & 37 & 0.93 & 0.93 & 0.90 & 0.79 & 0.86 & 0.89 & 0.93 & 0.93 & 0.88\\
pistol   & 38 & 0.68 & 0.68	& 0.71 & 0.78 & 0.65 & 0.72 & 0.78 & 0.65 & 0.72 \\
rocket  & 41 & 0.71	& 0.71 & 0.71 & 0.65 & 0.60 & 0.68 & 0.69 & 0.63	& 0.66\\
skateboard  & 45 & 0.84	& 0.86 & 0.87 & 0.88 & 0.81	& 0.84 & 0.87 & 0.89 & 0.79 \\
table   & 47 & 0.83	& 0.80	& 0.84 & 0.78 & 0.60 & 0.88 & 0.18 & 0.79 & 0.54 \\
\bottomrule
\end{tabular}
\label{tab_part}
\end{table*}

\subsection{Unsupervised Part Segmentation Using Hidden Semantics}
\begin{figure}[t]
\begin{center}
        \includegraphics[width=\linewidth]{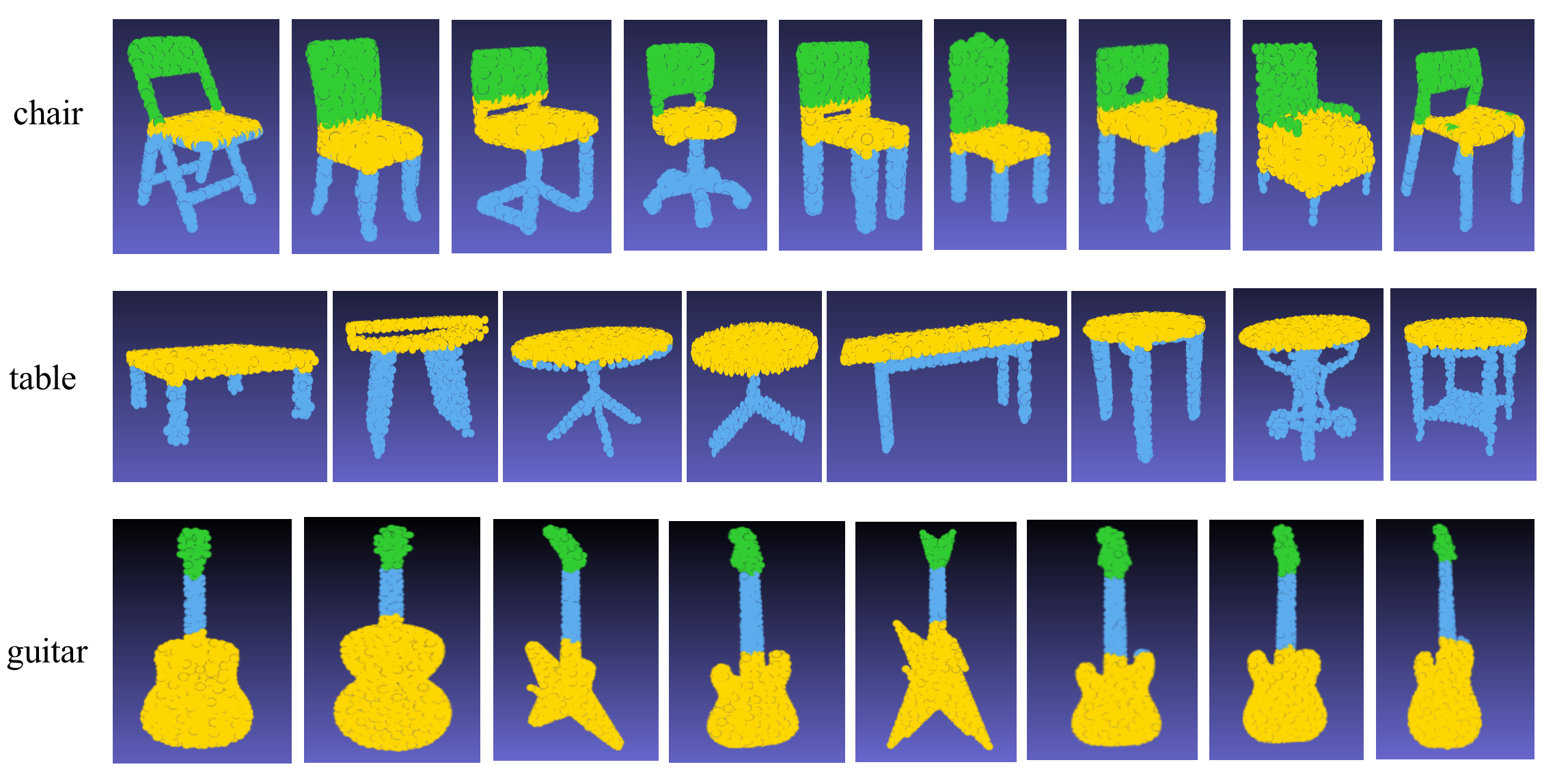}
        \caption{Unsupervised part segmentation using the hidden sementics of PointNet++ on ModelNet40. The results of chair and table rely on the hidden semantics of SA1-layer3, and the results of guitar rely on the SA3-layer1 and SA3-layer2. }	
        \label{ex_seg}
\end{center} 
\end{figure}

We achieve unsupervised part segmentation using the plane-level hidden semantics of PointNet++ on ModelNet40. After getting the salient region, composed by the salient points, the bounding box of the salient segmentation is attained. Utilizing the bounding box, as a 3D-mask, it easy to divide the object into different parts, as shown in Figure \ref{ex_seg}. 

\begin{figure}[!t]
\begin{center}
        \includegraphics[width=\linewidth]{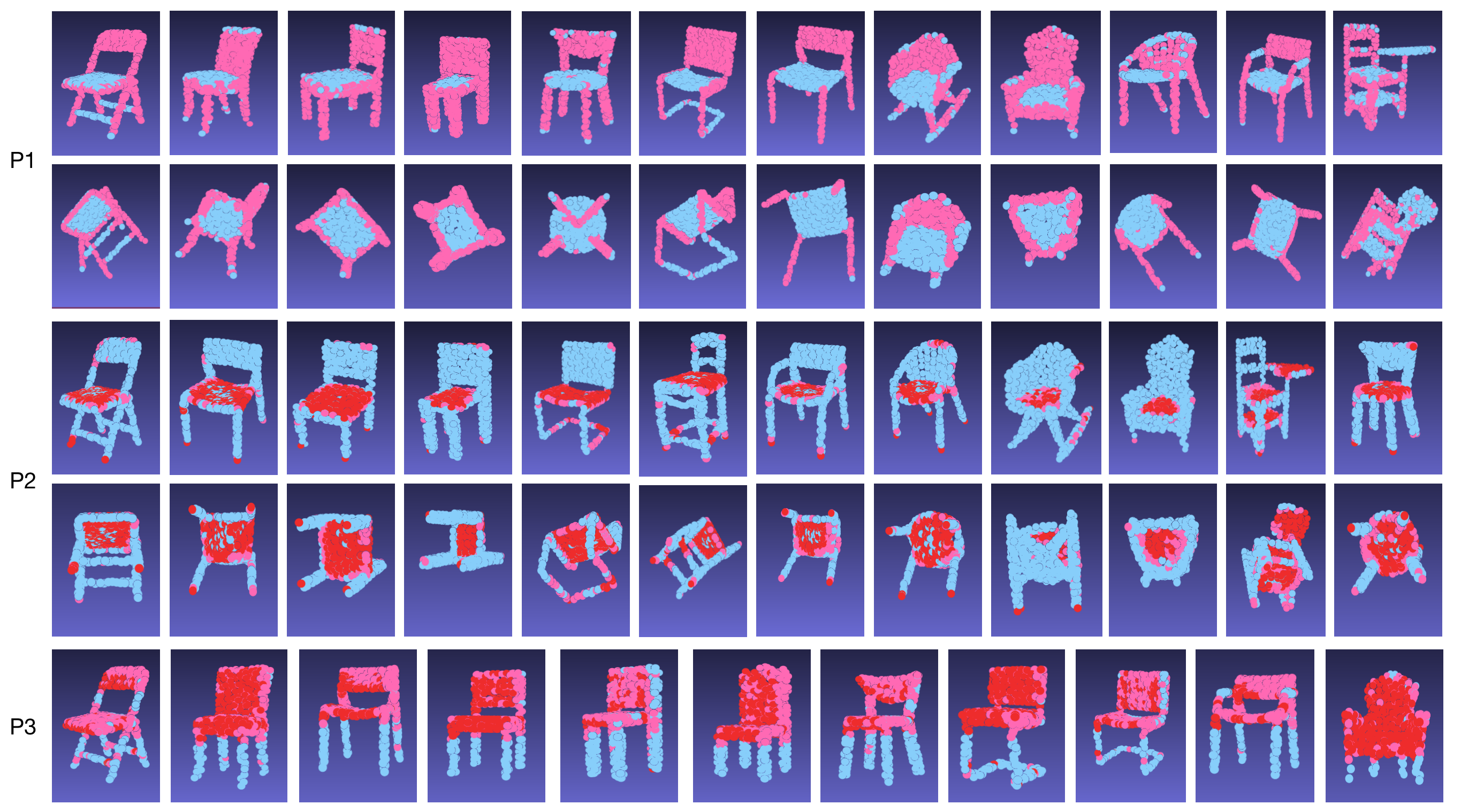}
        \caption{The result of part detector in the intermediate layers of PointNet++. The pattern $P1$ is the result of SA1-layer1, and $P2$ is the result of SA1-layer3. The pattern $P3$ comes from SA3-layer1. The red and pink regions are salient. }	
        \label{ex_detector}
\end{center} 
\end{figure}

We also conduct the part segmentation experiment on the ShapeNet part dataset. We trained the PointNet++ models without the segmentation lables, and feed the category labels only. Then we get the salient map as the part detector, as shown in Figure \ref{ex_detector}, of each layers by interpreting the hidden semantics of the network using our proposed Relevance Flow. Using the part detector of each layer, it is feasible to realize part segmentation. We calculate the mIoU of each category with the original sementation labels, as shown in Table \ref{tab_segmantation}. Compared with the state-of-the-art in point cloud part segmentation, our method get lowwer mIoU. The reason is that our method relies on the quality of the hidden semantics, however, without any part-level supervised learning, obtained from classification network, the hidden semantics is not a truly segmentation mask, as shown in Figure \ref{network}. The better effect requires subsequent improvements and restrictions on the training process. 

\begin{table}[t]
  \caption{The mIoU of the Part Segmentation on ShapeNet Part Dataset. Our performance without the segmentation labels, only rely on the hidden semantics in the trained  classification network. }
  \label{tab_segmantation}
  \begin{tabular}{lll}
    \toprule
    Method & Manner & mIoU\\
    \midrule
    PointNet & supervised   & 0.837\\
    PointNet++ & supervised & 0.851\\
    CurveNet & supervised   & 0.866 \\
    Ours & unsupervised & 0.713 \\
  \bottomrule
\end{tabular}
\end{table}

\subsection{Adversarial Attack Based on Hidden Semantics}

\begin{table*}
  \caption{The Success Rate and Time of Attack.}
  \label{tab_region_all}
  \begin{tabular}{ccccccc}
    \toprule
    \multirow{2}{*}{Network} & \multicolumn{2}{c}{PointNet}
	& \multicolumn{2}{c}{PointNet++} & \multicolumn{2}{c}{PointConv}\\
	\cline{2-7}
	& Ours & Random & Ours & Random & Ours & Random
	\\
    \midrule
    Mean Success Rate(\%)  & 28.7 & 17.2 & 11.3 & 7.8 & 20.4 & 15.0 \\
    Mean Time(s) 	& 0.19 & 0.20 & 3.13 & 2.64 & 0.34 & 0.34\\
    \bottomrule
  \end{tabular}
\end{table*}

\textbf{Success Rate and Time of Attack} We conduct adversarial attack experiments on PointNet, PointNet++, and PointConv networks with dataset ModelNet10 and ModelNet40. 

The result is show in Table \ref{tab_region_all}, 'Ours' means using our method to pick up the centers of attacked regions, and 'Random' means picking up the centers of attacked regions randomly. 
It shows that 'Ours' has higher success rate and lower time time consume than 'Random'. 
In terms of attack success rate, PointNet has the highest attack success rate, followed by PointConv, and PointNet++ is the lowest. It means that PointNet++ and PointConv networks are more robust and more resistant to avoid attack than PointNet. 
In terms of attack time, the time to attack PointNet is the shortest, followed by PointConv, and PointNet++ is the longest, because PointNet++ has the highest network complexity and the longest forward propagation time.

\begin{table}[h]
	\centering
	\caption{Attack success rate and time in PointNet on ModelNet10.}
	\begin{tabular}{cccccc} 
		\toprule
		\multirow{2}{*}{Class} & \multirow{2}{*}{Accuracy}
		& \multicolumn{2}{c}{Ours} & \multicolumn{2}{c}{Random}\\
		\cline{3-6}
		& & Rate(\%) & Time(s) & Rate(\%) & Time(s) \\
		\midrule
		bathtub              & 90 & 78.8 & 0.23 	& 6.7	& 0.09 \\
		bed                  & 75 & 20.5 & 0.20 	& 16.9 	& 0.20 \\
		chair                & 94 & 8.5	 & 0.18	 	& 8.5	& 0.24 \\
		desk                 & 91 & 10.6 & 0.19	 	& 5.8	& 0.21 \\
		dresser              & 55 & 69.2 & 0.19     & 57.7	& 0.23\\
		monitor              & 100 & 0.0   & 0.24	    & 0.0		& 0.21 \\
		night \space stand   & 80 & 60.5   & 0.18       & 42.0	    & 0.22\\
		sofa                 & 99 & 0.0	 & 0.18	 	& 1.0		& 0.21\\
		table                & 78 & 38.92 & 0.23	 	& 32.9	& 0.18\\
		toilet               & 88 & 0.0	 & 0.07	 	& 0.0		& 0.23\\
		Mean                 & 86 & 28.7 &	0.19	& 17.2	& 0.20\\
		\hline
	\end{tabular}
	\label{tab_pn_region}
\end{table}

Table \ref{tab_pn_region} shows the attack success rate and time in each of the class on PointNet. The number of regions is 5, and the number of neighbor points is 40. The 'Accuracy' refers to the classification accuracy of target category. 'Ours' means using our method to pick up the centers of attacked regions, and 'Random' means picking up the centers of attacked regions randomly. "Rate" means the success rate of adversarial attack, and 'time' refers to the time to generate an attack sample and attack the network at once. Table \ref{tab_pn_region} also shows the better performance of our method in attack success rate and efficiency.

\textbf{The Impact of the Set of Attack Region} We explore the impact of the number of attack region and the range of neighborhood on the attack result of PointNet network.

\begin{figure}[h]
\begin{center}
        \includegraphics[width=0.45\linewidth]{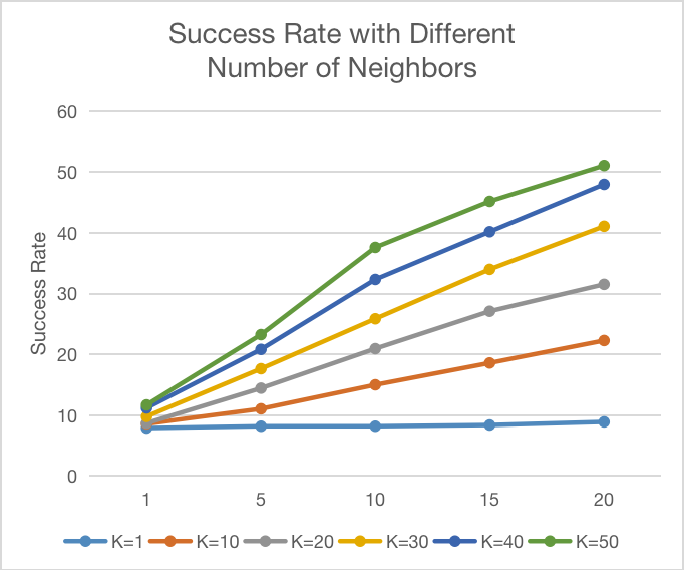}
        \includegraphics[width=0.45\linewidth]{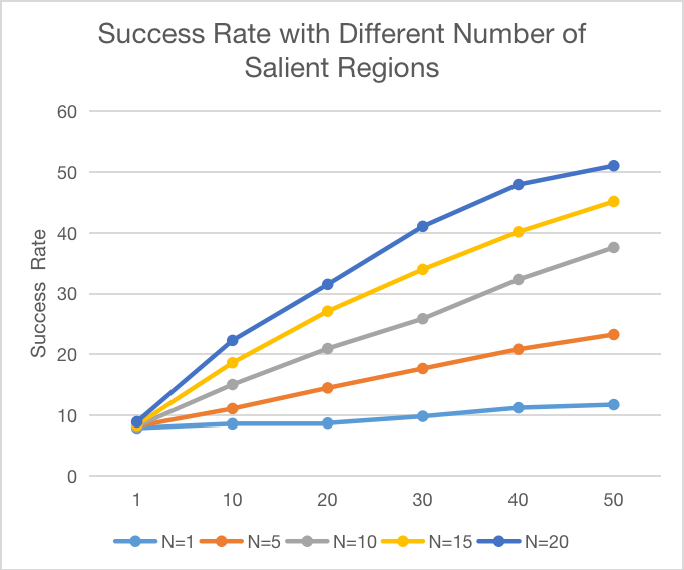}
        \caption{The success rate varies with different number of attack region and neighborhood.}	
        \label{chap5_set_attack1}
\end{center} 
\end{figure}

Figure \ref{chap5_set_attack1}. shows the success rate varies with different attack region number $N$ with the value: 1, 5, 10, 15, and 20. The abscissa in the figure represents number of regions, and the ordinate means the value of success rate, and different colors represent with different neighborhood range. 

In general, The Figure \ref{chap5_set_attack1} shows that the attack success rate of PointNet increases with the increase of the number of attack region and the range of neighborhood.

\section{Conclusion}
In this paper, the Relevance Flow, an interpretation method for 3D point cloud classification neural network, is proposed to interpret hidden semantics of the intermediate layers, and the saliency map is used to show the hidden semantics. Our proposed method reveals a explicit hidden semantics of different layers in PointNet, PointNet++ and PointConv frameworks. Particularly, we reveal the plane and part level hidden semantics in the intermediate layers of 3D point cloud classification neural network on the ModelNet40 and ShapeNet datasets. We use the normals and the IoU of salient points to evaluate the consistency of hidden semantics at both plane and part level. 
Beside, we part the segmentation of 3D point cloud using the hidden semantics in the intermediate layers with unsupervised manner, and generate the adversarial samples to attack the 3D point cloud DNNs.
Experiment shows that the plane-level and part-level hidden semantics can be retrofitted for unsupervised point cloud part segmentation with the trained classification neural network, and the attack success rate of PointNet is the highest, followed by PointConv and PointNet++.




\bibliographystyle{ACM-Reference-Format}
\bibliography{sample-base}

\appendix

\end{document}